\documentclass{article}

\PassOptionsToPackage{numbers, compress}{natbib}
\usepackage[final]{neurips_2019}

\usepackage[utf8]{inputenc} 
\usepackage[T1]{fontenc}    
\usepackage{hyperref}       
\usepackage{url}            
\usepackage{booktabs}       
\usepackage{amsfonts}       
\usepackage{nicefrac}       
\usepackage{microtype}      
\usepackage{algorithm}
\usepackage[]{algpseudocode}
\usepackage{shortcuts}
\usepackage{chemfig}
\usepackage{bm}
\usepackage{color}
\usepackage{amsmath,amsthm,amssymb,parskip,setspace,tabularx, wrapfig,float,enumerate} 
\usepackage{tikz, setspace, subfig, grffile, graphicx}
\usepackage{subfig}
\usepackage{paralist}
\usetikzlibrary{bayesnet}
\usepackage{float}

\title{Neural Message Passing on High Order Paths}

\begin{document}

\author{%
  Daniel Flam-Shepherd $^{1,4 *}$, Tony Wu $^{1 *}$, Pascal Friederich $^{1,2}$, Alan Aspuru-Guzik$^{1,3,4}$\\
  University of Toronto$^1$, Karlsruhe Institute of Technology$^2$, CIFAR$^3$, Vector Institute$^4$
}
\maketitle

\begin{abstract}
    Graph neural network have achieved impressive results in predicting molecular properties, 
    but they do not directly account for local and hidden structures in the graph
    such as functional groups and molecular geometry. 
    At each propagation step, GNNs aggregate only over first order neighbours, ignoring important 
    information contained in subsequent neighbours as well as the relationships between those higher order connections. 
    In this work, we generalize graph neural nets to pass messages and aggregate across 
    higher order paths. This allows for information to propagate over various levels and substructures of the graph.
    We demonstrate our model on a few tasks in molecular property prediction.  
\end{abstract}

\section{Introduction and Motivation}

Graph Neural Networks (GNNs) are a powerful tool for representation learning across different domains involving relational data 
such as molecules \cite{DuvMacetal15nfp} or social and biological networks \cite{hamilton2017inductive}. 
These models learn node embeddings in a message passing framework \cite{gilmer2017neural} 
by passing and aggregating node and edge feature information across the graph using neural networks.
The learned node representations can then be used for any downstream procedure such as node or graph classification or regression. 
In particular, GNNs have been used to drastically reduce the computation time for predicting molecular properties \cite{gilmer2017neural}.

However, current GNN models still suffer from limitations as they only propagate information across neighbouring edges and, 
after propagation, use simple pooling of final node embeddings \cite{DuvMacetal15nfp, li2015gated}. 
This means that, in most models, nodes only learn about the larger neighbourhood surrounding them over many propagation steps. 
This makes it difficult for GNNs to learn higher order graph structure and impossible to learn in a single propagation layer.
However, such long range correlations are important for many domains, in particular, 
when learning chemical properties that depend on rings, branches, functional groups or molecular geometry. 

The only way to directly account for higher order graph properties is to pass messages over additional neighbours in every propagation layer of the GNN. 
Notice how much larger the neighbourhood of the atom gets when you consider second and third order neighbours (Figure 1). 
This work focuses on generalizing message passing neural networks to accomplish this. 

\begin{figure}[h]
\begin{centering}
\includegraphics[width=.9\textwidth]{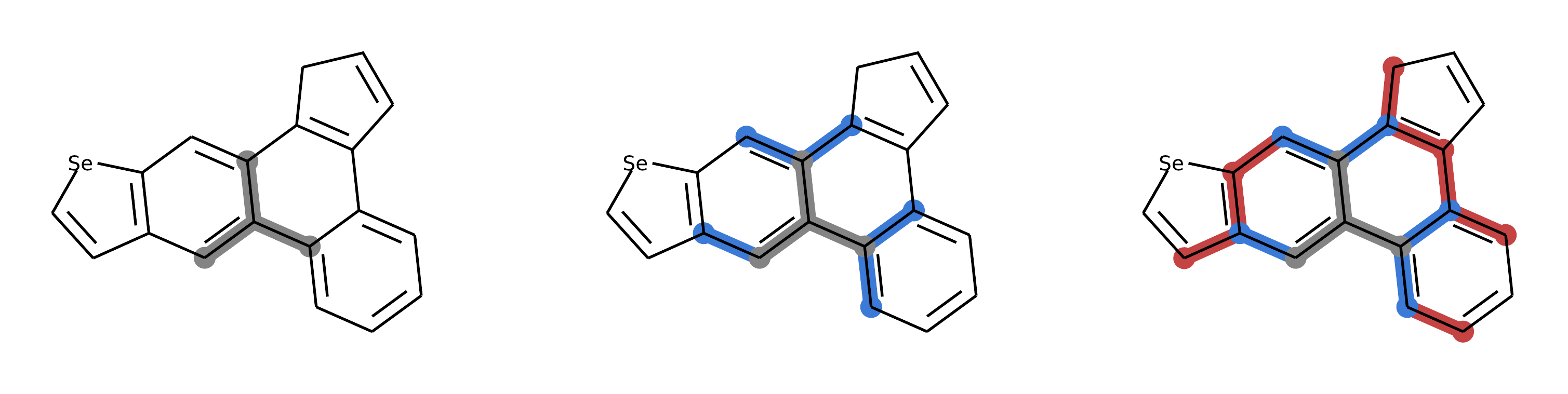} 

a) \hspace{3.5cm} b) \hspace{3.5cm} c) 
\caption{the neighbourhood of one atom comprised of 
a) $1^{st}$ b) $2^{nd}$ and c) $3^{rd}$ order neighbours}
\end{centering}
\end{figure}\label{fig:2Dgraph3Dmol}

\newpage 

 \subsection{Motivations}

There are many factors pertaining to molecular graphs that motivate the development of our model.  
In this section we discuss, in more depth, the limitations of GNNs with respect to specific aspects of molecules that motivate our model. 
These include molecular substructures like rings and functional groups, molecular geometry as characterized by internal coordinates as well as stereochemistry. 

\begin{wrapfigure}{r}{3.5cm}
\begin{centering}
\includegraphics[width=.2\textwidth]{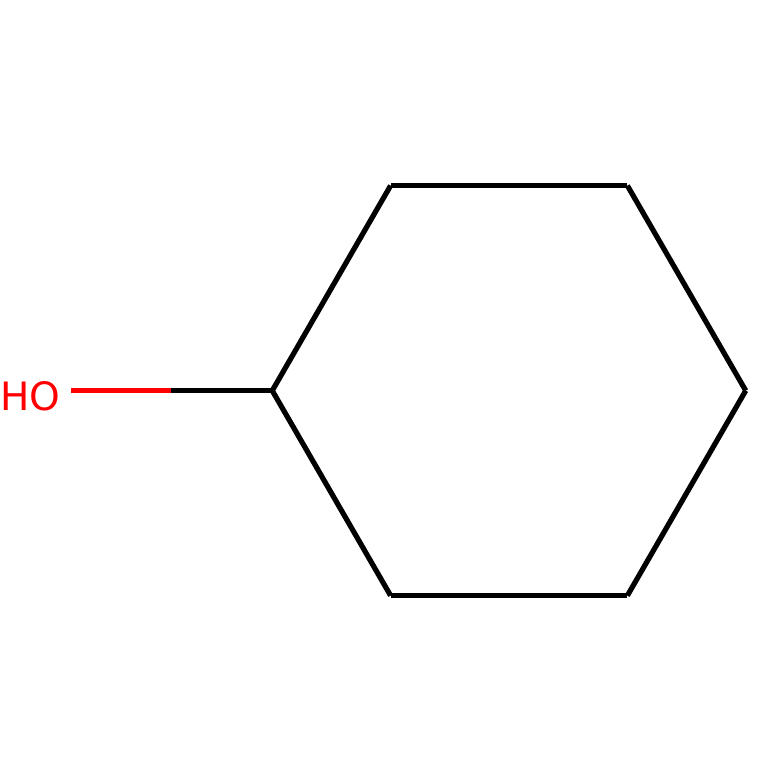} 
\caption{Cyclohexanol}
\end{centering}
\end{wrapfigure}\label{fig:2Dgraph3Dmol}

\textbf{molecular substructures } play an important role in determining molecular properties for example functional groups are responsible for the chemical reactions a molecule undergoes. 
By only aggregating over neighbours, GNN cannot learn about these larger substructures in a single propagation layer. 
On the other hand, by passing messages over larger neighbourhoods, in every layer we could directly learn about these structures. 
Furthermore, We could directly indicate if the path that a message is traveling on contains a simple functional group like alcohol (ROH) or passes through a larger functional group. 
For example, in figure 2, atoms in the neighbourhood of OH could receive messages of length two or more
indicating that an alcohol group is in their neighbourhood.

\textbf{molecular geometry} is the three dimensional arrangement of atoms in a molecule and influences several properties, including the reactivity, polarity and biological activity of the molecule.  
An important application of GNNs is predicting quantum mechanical properties of molecules, which are heavily dependent on the geometry of the molecule. 
The 3D configuration of a molecule can be fully specified by 1) bond lengths -- the distance between two bonded atoms, 2) bond angles -- the angle formed between three neighbouring atoms, and 3) dihedral angles between four consecutive atoms.
In fact the potential energy is typically modeled as a sum of terms involving each of these three.   
Current GNN approaches to quantum chemistry incorporate neighbouring geometry by using 
bond distances as edge features \cite{gilmer2017neural}, but do not directly account for the relative orientation of neighbouring atoms and bonds --
a framework that could do so would be advantageous. 

\begin{wrapfigure}{r}{5cm}
\begin{centering}
\includegraphics[width=.35\textwidth]{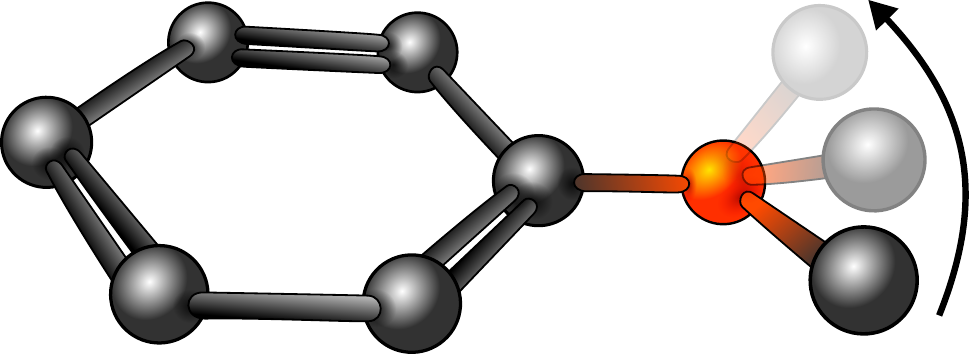} 
\caption{rotation of a bond}
\end{centering}
\end{wrapfigure}\label{fig:2Dgraph3Dmol}

\textbf{stereochemistry} involves the relative spatial arrangement of atoms in molecules, specifically, stereoisomers--
which are molecules with the same discrete graph but different three-dimensional orientation of atoms. 
For example, enantomers-- mirror images of molecules and cis-trans isomers, that only differ through the rotation of a functional group.
Even if they use interatomic distances as edge features, 
GNNs will have limited ability to distinguish stereoisomers, 
since these molecules only differ through the relative orientation of atoms.
In general, at every propagation step, 
GNNs should learn representations over each node's extended neighbourhood
to encode the relationships between nodes in that neighbourhood.

\subsection{Approach and Contributions} 
We generalize MPNNs to aggregate across larger neighbourhoods by passing messages along simple paths of higher order neighbours. We describe the general framework in Section 3. 
We experiment with various molecular property prediction task and a node classification task in citation networks. Our specific contributions are two-fold

\begin{itemize}
    \item we devise a simple extension to any message passing neural network to learn representations over larger node neighbourhoods within each propagation layer by simply augmenting the message function to aggregate over additional neighbours. 
    \item By summing over additional neighbours we enable the use of path features such as bond angles for paths of length two and dihedral angles for paths of length three and thus encoding the full molecular geometry and orientation, 
          so that MPNNs can distinguish isomers.  
\end{itemize}

\newpage

\section{Related Work and Background}

\subsection{Background} 
\textbf{Message passing neural networks} operate on graphs $G$ with $n$ nodes each with feature vector $\B x_v \in \R ^f$ 
that specify what kind of atom the node is, among other possible features. 
There are $n\times n $ edge feature vectors $\B e_{vw}\in \R ^e$ that specify what kind of bond type atoms  $ v, w $ have.
The forward pass has two phases, a message passing phase and a readout phase. 

The  message  passing  phase  runs  for $T$ propagation steps and is defined in terms of message functions $M_t$ and node update functions $U_t$.  
During the message passing phase, 
hidden states $\bm h^t_v$ at each node in the graph are updated based on messages $\bm m^{t+1}_v$ according to
\[ \bm m _v ^{t+1} = \sum _{w\in \N _v } M_t(\bm h_v ^t, \bm h_w^t, \bm e_{vw}) , 
\ \ \ \bm h^{t+1}_v = U_t(\bm h^{t}_v, \bm m^{t+1}_v), 
\ \ \ \B y = \texttt{Readout}(\{\bm h^T _v\}_{v\in G}) \] 
The message node $v$ receives aggregates over its neighbours $\N_v$, in this case, by simple summation.
We then readout predictions $\B y$ based on final node embeddings. 

\subsection{Related Work}

The first graph neural network model was proposed by \cite{scarselli2008graph} 
and many variants have been recently proposed \cite{li2015gated,velivckovic2017graph, kipf2016semi}. 
Our focus is on the general framework of neural message passing from \cite{gilmer2017neural}. 
We review relevant GNN models and their use in Molecular Deep learning in this section. 

\textbf{Molecular Deep Learning} Recently GNNs have superseded machine learning methods involving hand-crafted feature representation, on predicting molecular properties. 
For example, neural fingerprints generalizes standard molecular fingerprints with a differentiable one that achieves better predictive accuracy \cite{DuvMacetal15nfp}. 
Another model, SchNet \cite{schtt2017schnet} defines a continuous-filter convolutional neural network for modeling quantum interactions and achieves state of the art results. 

\textbf{Higher Order GNNs.}  
Recent work has generalized graph convolution networks (GCNs) \cite{kipf2016semi} to higher order structure by repeatedly mixing feature representations of neighbors
at various distances \cite{abu2019mixhop}, or casting GCNs into a general framework inspired by the path integral formulation of quantum mechanics \cite{ma2019pan}. Both of these works are based on powers of the adjacency matrix and do not account directly for the relationship between higher order neighbours. Another work \cite{morris2019weisfeiler} proposes k-dimensional GNNs in order to take higher order graph structures at multiple scales into account. 
GNNs and higher order GNNs do not incorporate the relationship between higher order neighbours, 
which would allow for features that are dependent on that relationship, namely 'path features'.

\textbf{path augmented transformer}
Another model based on the transformer architecture \cite{chen2019path} accounts for long range dependencies in molecular graphs by augmenting edge feature tensor to include some (shortest) path features like bond type, conjugacy, inter-atomic distance and ring membership.

\textbf{structured transformer} A few graph neural networks recently proposed have incorporated directional information. The first 
\cite{ingraham2019generative} builds a model for proteins that considers the local change in the coordinate system for each atom in the chain. 

\textbf{3D GCN} \cite{cho2014properties} build a three-dimensional graph convolutional network, 
for molecular properties and biochemical activities prediction using 3D molecular graph by augmenting the standard GCN layer with the relative atomic position vector. 

\textbf{directional message passing} \cite{Klicpera2020Directional} embeds the messages passed between atoms such that each message is associated with a direction in coordinate space and 
are rotationally equivariant since the associated directions rotate with the molecule. 
Their message passing scheme transforms messages based on the angle between them in order to encode direction.

\newpage 
\begin{figure}[H]
\centering
  \def\svgwidth{.9\linewidth}
    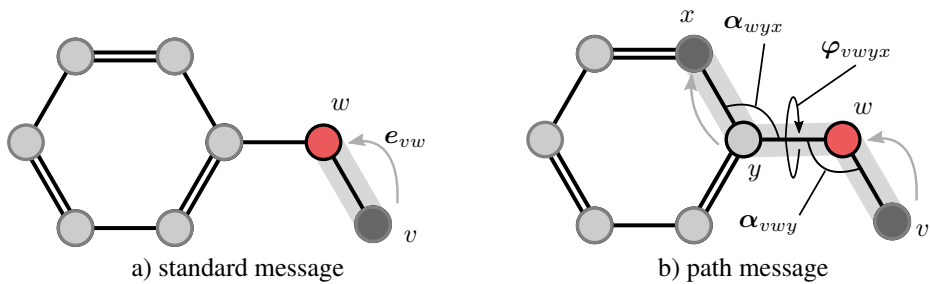

a) standard message \hspace{4cm} b) path message
\caption{Message function and path features for a) standard MPNN and b) MPNN passing messages on paths with length 3 in a molecule with path features involving molecular geometry}
 \end{figure} 

\section{Neural Message Passing on paths }

We extend the message passing framework by propagating information from every node's higher order neighbour instead of aggregating messages from only nearest neighbours.
The message passing phase is augmented such that hidden states $\bm h^t_v$ at each node in the graph are updated based on messages over all simple paths up to length $\ell$ from its neighbourhood: 
\begin{align}
\bm m_v ^{t+1} = \sum _{\bm p \in \mathcal{P}_\ell^v } M_t(\bm h_v ^t, \bm p ) = 
\sum _{v_1 \in \N _v } 
\sum _{\substack{v_2 \in \N _{v_1} \\ v_2 \neq v} } \cdots
\sum _{\substack{v_\ell \in \N_{v_{\ell -1}} \\ v_\ell \neq v_{\ell-2} , \dots , v}  } M_t(\bm h _v ^t, \bm p_{v_1 \to v_\ell} )
\end{align}

where we define $\bm p$ to be a path in $\mathcal{P_\ell}^v $, which is the set of all simple paths starting from node $v$ with length $\ell$ and $\bm p_{v_1 \to v_\ell}$ to be path features along path $\bm p$ from node $v_1$ to node $v_\ell$. We only sum over simple paths, excluding loops and multiple inclusions of the same node.

\subsection{Path features}

For graphs with a large number of nodes and edges, passing messages along paths becomes very expensive and, as in GraphSage \cite{hamilton2017inductive}, sampling a subset of paths of higher order neighbours is necessary.
However, for molecules, where the number of neighbours is usually $\leq 4$ this is not necessary.  
Furthermore, one can include domain specific path features in the message function. We describe two examples of these path features below  

\textbf{molecular substructures } we can incorporate whether the path travels through a molecular substructure by considering paths of at least length 2, 
where we have a message function that sums over 2 neighbouring atoms $ v \to w \to y $. Along with their node and edge features, the possible path features 
include ring features - ie one hot indication if any atoms are in (specific) rings as well as if the path is a functional group (ROH) or within a larger functional group.  
\begin{align}
\bm m _v ^{t+1} = \sum _{w\in \N _v } \sum _{\substack{y \in \N_w \\ y\neq v}}  M_t( \bm h_{v}^t, \bm p_{v\to y})   \ \  \ \ \
\bm p_{v\to y} = \begin{bmatrix} \bm h_w^t & \bm h_y ^t \\  \bm e_{vw} &  \bm p_{vy}  \end{bmatrix}
\end{align}

\textbf{molecular geometry } considering paths of length 3, where we have a message function that sums over 3 neighbouring atoms  $ v \to w \to y \to x $. 
Along paths of length three additional features include two bond angles $\bm \alpha_{vwy} \ \& \ \bm \alpha_{wyx} $ 
and the dihedral angle $ \bm \varphi _{vwyx}$ between the planes defined by the pairs of atoms $(v,w)$ and $(y,x)$.  
Effectively, messages passed over 3 consecutive neighbours contain information about the entire molecular geometry (see Figure 4).
\begin{align}
\bm m _v ^{t+1} = \sum _{w\in \N _v } \sum _{\substack{y \in \N_w \\ y\neq v}} \sum _{\substack{x \in \N_y \\ x \neq w,v }} M_t( \bm h_{v}^t, \bm p_{v\to x})   \ \  \ \ \
\bm p_{v\to x} = \begin{bmatrix} \bm h_w^t & \bm h_y ^t& \bm h_x ^t \\  \bm e_{vw} &  \bm e_{wy}& \bm e_{yx} \\ \bm \alpha_{vwy}&\bm \alpha_{wyx}& \bm \varphi _{vwyx} \end{bmatrix}
\end{align}

\begin{figure}[t]
\begin{centering}
\begin{tabular}{ lccc } 
Dataset & QM8 & ESOl & CEP\\ 
Units & MAE in eV ($\times 10^{-3}$) & RMSE in log Mol/L  & Percent\\ 
 \toprule
 neural fingerprint \cite{DuvMacetal15nfp}  & 13.80 $\pm$ 0.11 & 0.52 $\pm$ 0.07 & 1.43 $\pm$ 0.09 \\
 message passing NN (ours)  & 11.30 $\pm$ 0.31 & 0.47 $\pm$ 0.03 & 1.37 $\pm$ 0.09 \\
 molNet \cite{wu2018moleculenet}              & 10.80 $\pm$ 0.30 & 0.58 $\pm$ --   &  -- \\
 path transformer \cite{chen2019path}    & 10.20 $\pm$ 0.30 & 0.55 $\pm$ 0.06 &  --  \\
 \toprule
 \textbf{Path MPNN} & \textbf{8.70 $\pm$ 0.06} & \textbf{0.41 $\pm$ 0.02} & \textbf{1.23 $\pm$ 0.08} \\
 \toprule
\end{tabular}

\vspace{.25cm}
 Table 1 : Mean and std error predictive accuracy on various dataset and baselines
\vspace{.25cm}

\includegraphics[width=.25\textwidth]{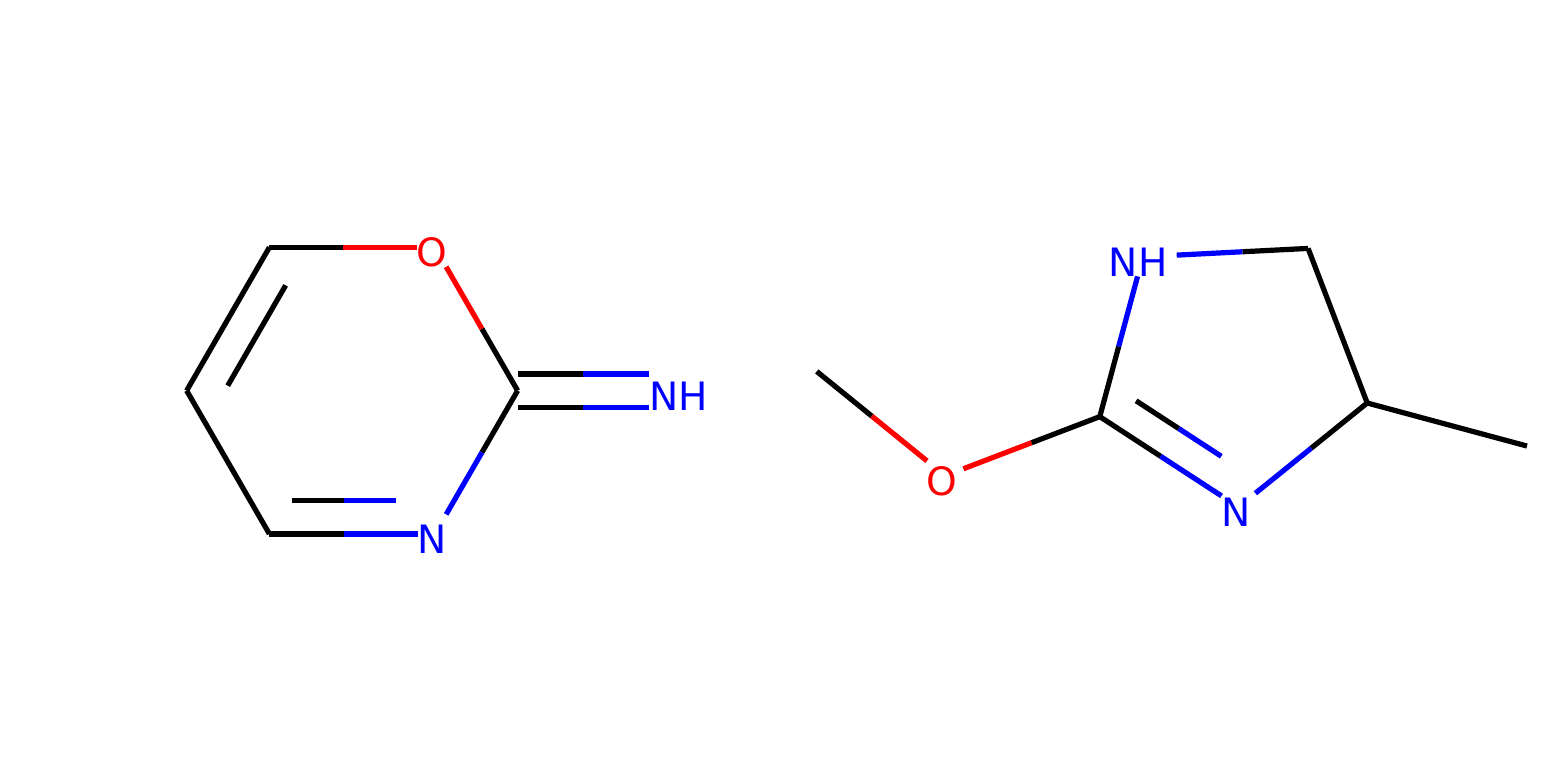} \vline 
\includegraphics[width=.25\textwidth]{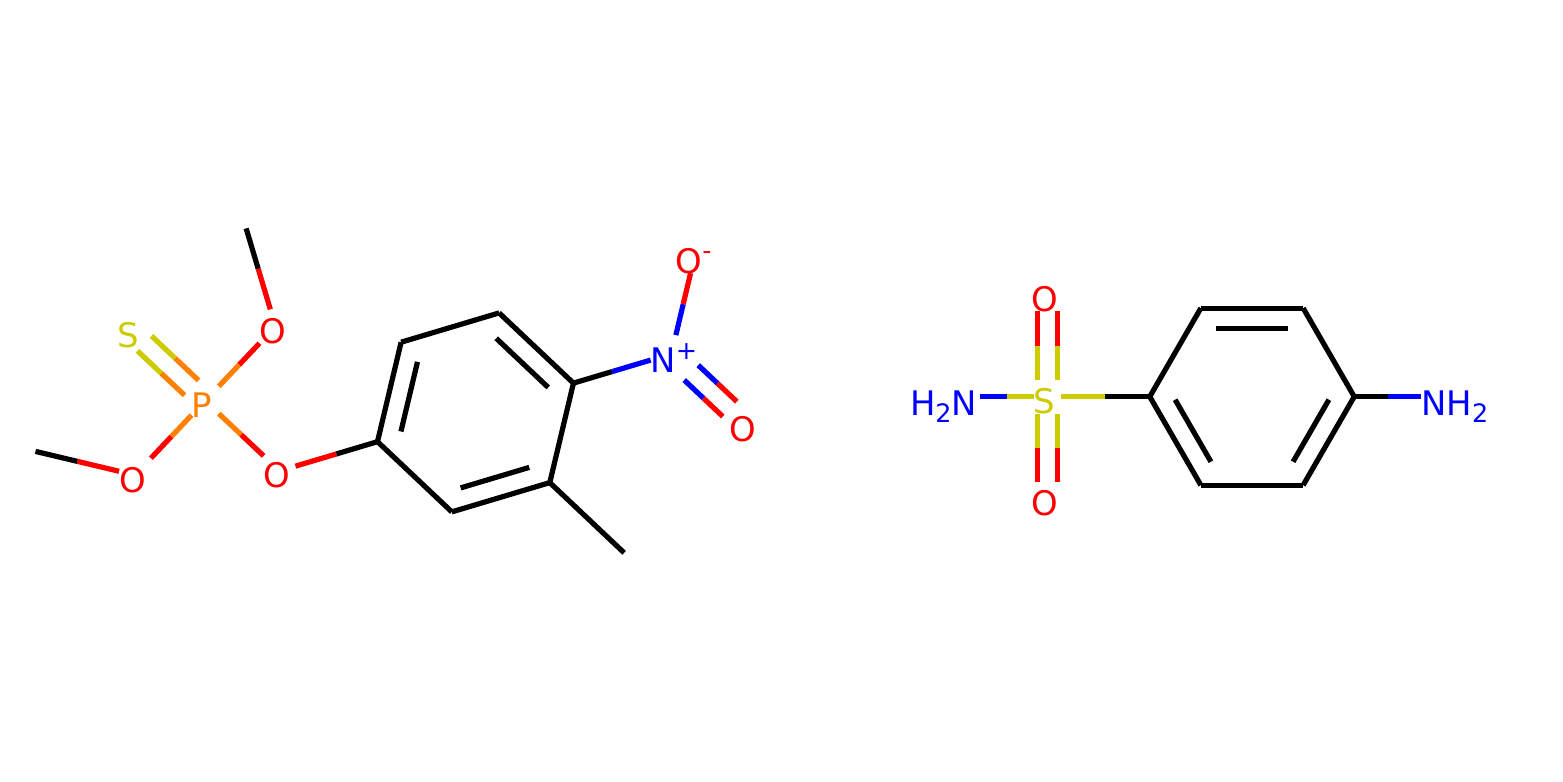} \vline 
\includegraphics[width=.25\textwidth]{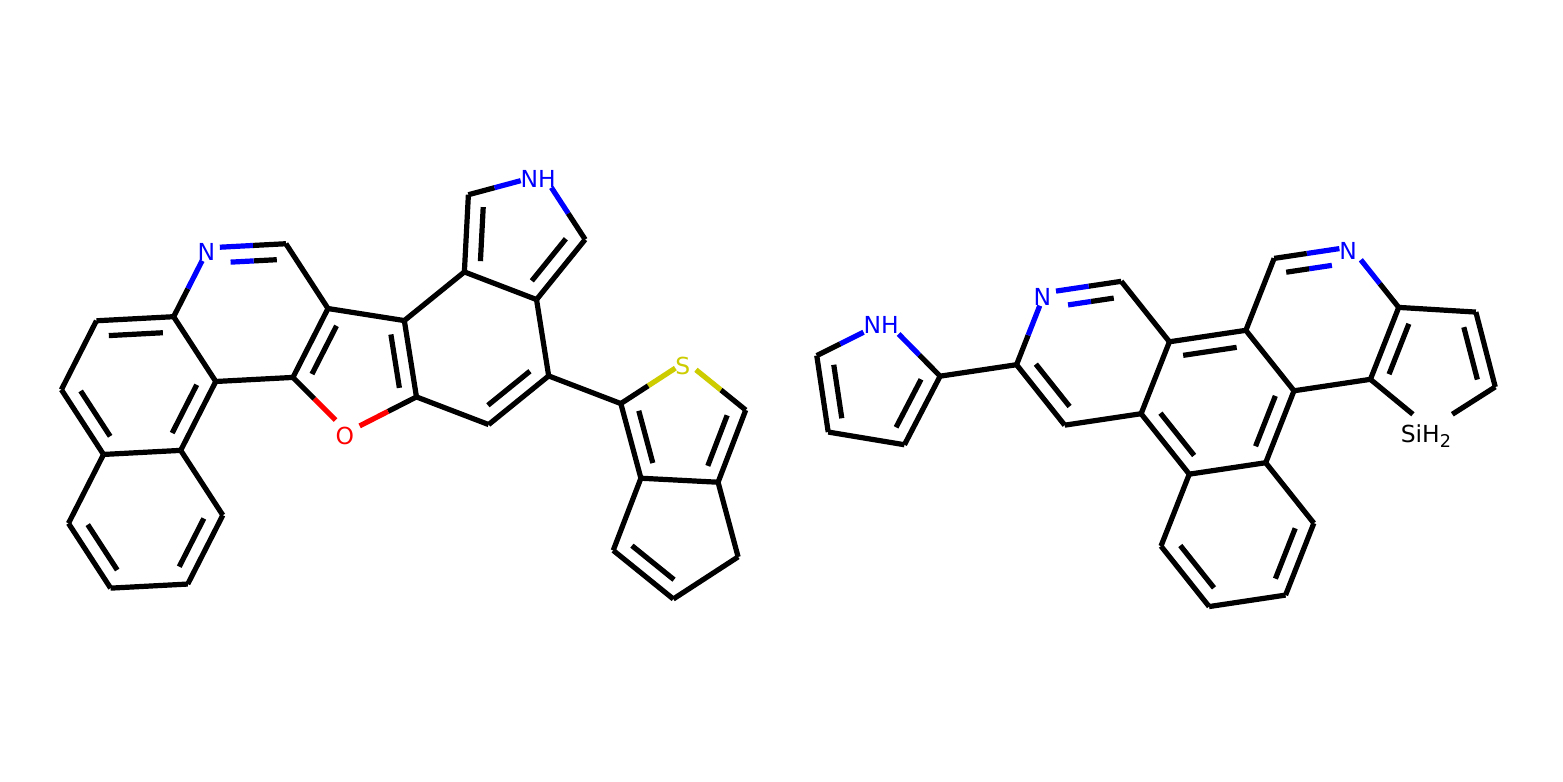} 

a) QM8 \hspace{2cm} a) ESOL \hspace{3cm} a) CEP 

\caption{molecules from the datasets considered}
\end{centering}
\end{figure}\label{fig:2Dgraph3Dmol}

\section{Experiments}

\subsection{Datasets} \ \  We compare the performance of our model against a few baselines on a variety of molecular property prediction tasks involving different datasets. These tasks include :
\begin{itemize}
    \item ESOL: \cite{delaney2004esol} predicting the aqueous solubility of 1144 molecules.
    \item QM8 : \cite{ruddigkeit2012enumeration} predicting 16 electronic spectra values calculated using density functional theory for 21786 organic molecules that have 8 or less heavy atoms (CON and F)  
    \item CEP : the photovoltaic efficiency of 20000  organic molecules from The Harvard Clean Energy Project \cite{hachmann2011harvard} 
\end{itemize}

\subsection{Model design}

We use the following basic MPNN model that is augmented along the lines of section three in order to pass messages over paths (Path MPNN). 
\begin{align*}
    \bm m_v^{t+1} = \underset{ w \in \N_v   }{\texttt{Attention}} \  M_t(\bm h_v ^t, \bm h_w^t, \bm e_{vw}) , \ \ \ 
     U_t = \sigma ( [\bm h_v^t, \bm m_v^t ]    ) 
     \ \ \  \B y = \underset{v\in G}{\texttt{Set2Set}}(\{\bm h^T_v, \bm x_v\} )
\end{align*} 
this uses graph attention \cite{velivckovic2017graph} as an aggregation method and the message function from the interaction networks model in \cite{battaglia2016interaction}, 
which is a simple concatenation of node and edge features.  
The node update function concatenates incoming messages with the current node state and feeds it through a dense layer. 
After propagation through message passing layers, we use the set2set model \cite{vinyals2015order}
as the readout function to combine the node hidden features into a fix-sized hidden vector.  For QM8 we pass messages over paths of length three and use path features for molecular geometry as specified in equation (3). 
For ESOL and CEP we pass messages over paths of length two and use path features for molecular substructures as specified by equation (2)
The models are trained using root mean squared error (RMSE) for loss. Model evaluation is done using mean absolute error (MAE) of the molecular properties in the QM8 dataset, RMSE for ESOL and percent for CEP.

\subsection{Results}

\textbf{Baselines} We use the top performing model from Molecule Net \cite{wu2018moleculenet} (Molnet) for each dataset. 
We also benchmark with the differentiable version of circular fingerprints from \cite{DuvMacetal15nfp} (neural fingerprints). 
To highlight the importance of path features, we also compared the performance of the (MPNN) model we used without passing messages on paths.
The last benchmark is the Path-Augmented Graph Transformer Network) (PAGTN) since this model is similarly built to model longer-range dependencies in molecular graphs.
As can be seen in Table 1, for QM8, ESOL and CEP, passing messages over paths leads to a substantial improvement in predictive accuracy.

\section{Comparison with other Higher Order GNNs}

In a separate experiment, we compare the path MPNN with other GNNs that use higher order neighbours and do not use edge features. 
We consider a standard task of semi-supervised node classification with the CORA dataset.

\subsection{The dataset} It contains sparse bag-of-words feature vectors for each document and a list
of citation links between documents which we use as undirected edges in the adjacency matrix. 
Each document has a class label. 
Altogether, the network has 2,708 nodes and 5,429 edges with 7 classes and 1,433 features. 

\begin{wraptable}{r}{6.0cm}
\begin{tabular}{ lc } 
Model & Test accuracy \\ 
 \toprule
 GCN \cite{kipf2016semi} & 81.5 \\
 MixHop \cite{abu2019mixhop} & 81.9 \\
 PAN \cite{ma2019pan} & 82.0 \\ 
 \toprule
\textbf{Path GCN} & \textbf{82.4} \\
\end{tabular}
\end{wraptable}

\subsection{Model} We use the experimental setup of \cite{kipf2016semi}. We sum over paths of length 3 while uniformly sampling a single second order and third order neighbour. 
Our base MPNN is a GCN \cite{kipf2016semi} that has message function 
$$  \bm m _v ^{t+1} = \sum _{w\in \N _v } \hat {\B A}_{vw} \bm h ^t_v , \ \ U_t = \sigma ( \bm m_v^t ) $$ 
where $\sigma$ is a dense layer with sigmoid activation. For a citation network the path features are just the node features and edge features connecting 
$v$ to nodes that are $\ell $ nodes away, {\it i.e.} 
\begin{align*}
    \bm p_{v_1 \to v_\ell} = \{ \bm h_{v_1}^t, \bm e_{vv_1}, \dots , \bm h_{v_\ell}^t, \bm e_{v_{\ell}v_{\ell -1}} \}
\end{align*}

\subsection{Results} We compare with two other higher order GCN variants: Mixhop \cite{abu2019mixhop} and PAN \cite{ma2019pan}: Path integral graph convolution -- both use powers of the adjacency to aggregate GCN layers of higher order neighbours. 
From the results table above our model achieve similar accuracy to our baselines.

\section{Conclusion and Discussion}

\textbf{Limitations} In this work we only considered very simple message functions, in general, it is not straight forward to construct message function over paths. 
For example, the message function from \cite{gilmer2017neural}, maps edge features to a square matrix using a neural net-- incorporating more neighbours and their edge and path features into this kind of message function introduces many design challenges. 

We introduce a general GNN framework based on message passing over simple paths of higher order neighbours. 
This allows us to use path features in addition to node and edge features, which is very useful in molecular graphs, as many informative features are characterized by the paths between atoms.
We benchmarked our framework on molecular property prediction tasks and a node classification task in citation networks. 

\bibliographystyle{unsrt}
\bibliography{paper}


\end{document}